# Real Time American Sign Language Detection Using Yolo-v9

**Amna Imran**
ETIT-KIT, Germany

**Meghana Shashishekhara Hulikal**
ETIT-KIT, Germany

**Hamza A. A. Gardi**
IIIT at ETIT-KIT, Germany

## Abstract

This paper focuses on real-time American Sign Language Detection. YOLO is a convolutional neural network (CNN) based model, which was first released in 2015. In recent years, it gained popularity for its real-time detection capabilities. Our study specifically targets YOLO-v9 model, released in 2024. As the model is newly introduced, not much work has been done on it, especially not in Sign Language Detection. Our paper provides deep insight on how YOLO-v9 works and better than previous model.

## 1 Introduction

### 1.1 Need of Sign Language Detection

Sign language is a visual means of communicating through hand gestures, body movements and facial expressions. It is an efficient tool for those with hearing and speaking disabilities to connect to the world and communicate their thoughts and ideas. According to data from the World Health Organization in 2024, there are around 430 million people suffering from hearing loss around the world, including 34 million children. It is estimated that this number will exceed 700 million – or 1 in every 10 people – by 2050. (World Health Organization, 2024). (WHO, 2024). Sign Language is a crucial element in life of deaf-mute people, but it is still not commonly known by other people, which makes it difficult to communicate and go by in day-to-day life. In order to bridge this gap and connect deaf-mute with the rest of the world, with the help of deep learning, Sign Language Recognition (SLR) model has been introduced to revolutionize the way sign language is integrated into daily life, enhance communication, accessibility, and inclusively.

There are over 300 different sign languages used around the world. Examples include American Sign Language (ASL), British Sign Language (BSL), and French Sign Language (LSF). Each sign language has its own grammar and syntax. For the purposes of this paper, we will use American Sign Language throughout the paper. This paper introduces a state-of-the-art YOLO-v9 model to predict the gestures performed in real-time at high accuracy. The model is capable of recognizing hand gestures that can be deployed in any real-life situation as well as multiple gestures within one image.

### 1.2 YOLO - You Only Look Once

YOLO (You Only Look Once) is a popular and powerful object detection algorithm used in computer vision. It is designed to identify and locate objects within an image or video frame in real-time. Unlike traditional object detection methods that often involve multiple passes over the image, YOLO processes the entire image with a single neural network, making it extremely fast and efficient. In 2015, the real-time object detection system YOLO was published, and it rapidly grew in iterations, each building upon the previous version to address limitations and enhance performance, with the newest releases, YOLO-v9 and YOLO-v10(Wang et al., 2024a) in 2024. (Figure 1)

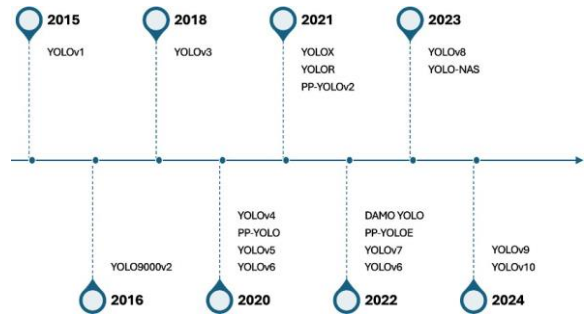

Figure 1: YOLO Evolution over the years

### 1.3 Related Work

In addition to the YOLO algorithm, several other great methods have been developed on object detection and image processing. Techniques such as R-CNN (Region-based Convolutional Neural Networks) (Girshick et al., 2014), Fast R-CNN (Girshick, 2015) and Faster R-CNN (Ren et al., 2017). These methods work on a two-stage process, where selective search generates region proposals, and convolutional neural networks classify and refine these regions. Another single-step technique, similar to YOLO is the Single-Shot Multi Box Detector (SSD) (Liu et al., 2016), which removes the need for a separate region proposal step making the model fast and efficient. Each method has its own unique trade-offs between time, performance, and complexity, catering to different application needs and computational constraints.

In recent years, YOLO (You Only Look Once) has become popular CNN method in Sign Language Recognition (Liu et al., 2024; Hsu and Lin, 2021; Katyayani et al., 2023). It is evident that YOLO provides high accuracy and performance in computer vision detection.

## 1.4 Object Detection Metrics

To determine and compare the predictive performance of different object detection models, standard quantitative metrics are used. The two most common evaluation metrics are Intersection over Union (IoU) and Average Precision (AP).

### 1.4.1 Average Precision

Average Precision is the area under the precision-recall curve. Precision is the ratio of true positive detection to the total number of detection (both true positives and false positives). It measures the accuracy of the positive predictions and given as;

$$Precision = \frac{TruePositive}{TruePositive + FalsePositive} \quad (1)$$

Whereas, Recall is the ratio of true positive detection to the total number of ground truth instances. It measures the ability of the model to find all relevant instances and given as;

$$Recall = \frac{TruePositive}{TruePositive + FalseNegative} \quad (2)$$

### 1.4.2 Intersection over Union

Many object detection algorithms use the intersection of union (IoU) function for the bounding box prediction. IoU is the ratio of the intersection area to the union area of the predicted bounding box and the ground truth bounding box. (Figure 2) (Terven et al., 2023)

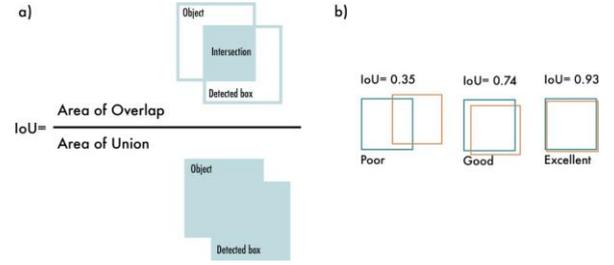

Figure 2: Intersection over Union (IoU). a) The IoU is calculated by dividing the intersection of the two boxes by the union of the boxes; b) examples of three different IoU values for different box locations.

## 1.5 How Does YOLO works?

YOLO is actually based on the idea of segmenting an image into a grid of cells, not smaller images. The image is split into a square grid of dimensions S×S. Each Grid cell predicts B bounding boxes and confidence score. The confidence score represents the model's confidence that an object is present in the predicted box, and is calculated as;

$$Confidence(C) = P(Object) * IOU_{pred}^{truth} \quad (3)$$

There are 5 predictions: x, y, w, h, and confidence inside each Bounding box B. The (x, y) coordinates corresponds to the center of the box relative to the boundary of the grid cell. The width and height are predicted relative to the whole image. Lastly, the confidence prediction represents the IOU between the predicted box and any ground truth box.

Each grid cell also predicts C conditional class probabilities, $P(Class_i|Object)$. These probabilities are conditioned on the grid cell containing an object. the model only predicts one set of class probabilities per grid cell, regardless of the number of boxes B. During test it multiply the conditional class probabilities and the individual box confidence predictions,

$$C = P(Class_i|Oobject) * P(Oobject) * IOU_{pred}^{truth} \quad (4)$$

$$C = P(Class_i) * IOU_{pred}^{truth} \quad (5)$$

which gives a class-specific confidence scores for each box. These scores encode both the probability of that class appearing in the box and how well the predicted box fits the object. (Redmon et al., 2016)

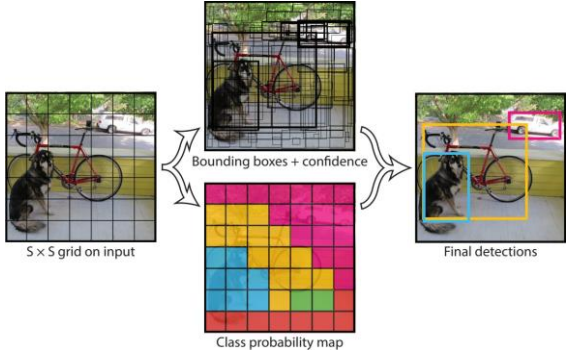

Figure 3: The model divides the image into an S × S grid and for each grid cell predicts bounding boxes, confidence for those boxes, and class probabilities. These predictions are encoded as an $(SxSx(B*5+C))$ tensor.

## 2 Dataset

In order to train and evaluate the Yolo-v9 model, the dataset used for this paper comprises of 26 classes correspond to 26 letters of American Sign Language: A, B, C, D, E, F, G, H, I, J, K, L, M, N, O, P, Q, R, S, T, U, V, W, X, Y, Z. There are a total of 1215 training images, 77 validation images and 42 test images. All images are resized to 640 x 640. The dataset contains both single object images and multiple object images.

## 3 Problem statement

In deep neural networks, as data flows through multiple layers, some of the original information gets lost. This is because each layer transforms the data, and these transformations can filter out certain details. This phenomenon is called 'information bottleneck'. Deep learning methods aim to optimize objective functions for accurate predictions close to the ground truth. The architecture design is crucial for acquiring sufficient information for prediction tasks. Existing methods often overlook the substantial information loss during the layer-by-layer feature extraction and spatial transformation of input data in deep networks. YOLOv9 (Wang et al., 2024b)addresses these issues of data loss in deep networks by specifically focusing on the information bottleneck and reversible functions.

### 3.1 Information Bottleneck Principle

The information bottleneck principle offers a new perspective on deep learning by viewing it as a

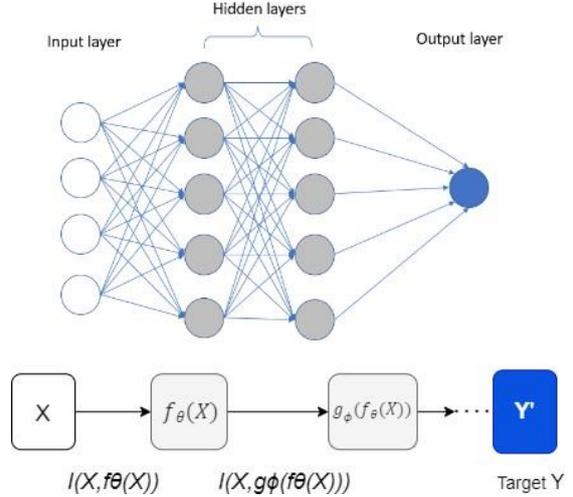

Figure 4: This diagram illustrates the process of information loss in a deep neural network as data $X$ passes through multiple transformation layers $f_\theta$ and $g_\phi$. The mutual information $I$ decreases at each layer. The network's output $Y'$ is compared to the target $Y$ using a loss function, which generates gradients for updating the network parameters.

problem of representation learning. According to this principle, each layer of a Deep neural network functions as a summary statistic retaining relevant information about the target output while discarding irrelevant details from the input. The mutual information I(X;Y) quantifies the relevant information that the input X contains about the output Y.(Sakamoto and Sato, 2024)

The loss of information during the feedforward process can result in biased gradient flows. These gradients are crucial for updating network parameters during training. If the gradients are unreliable due to information loss, the network may learn incorrect associations between inputs and targets, leading to poor performance and incorrect predictions.

One of the primary challenges in deep neural networks is the trade-off between depth and width. As networks become deeper, they are likely to suffer significant information loss. This phenomenon is highlighted by the inequality:

$$I(X, X) \geq I(X, f_\theta(X)) \geq I(X, g_\phi(f_\theta(X)))$$

where $I$ denote mutual information, and $f_\theta$ and $g_\phi$ are transformation functions parameterized by $\theta$ and $\phi$, respectively. This inequality shows that the mutual information between the input and the transformed data decreases as the data passes through

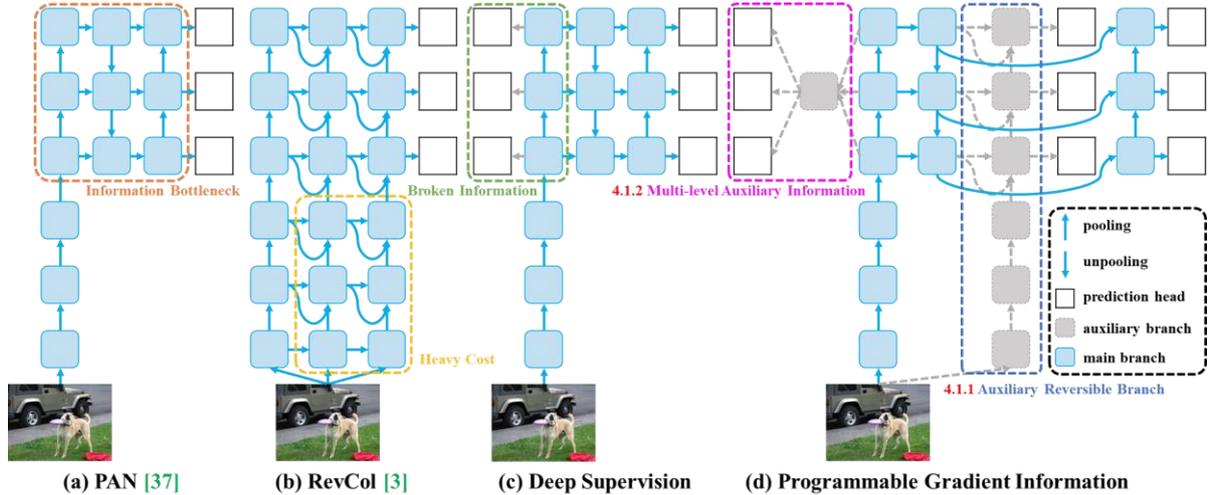

Figure 5: PGI and related network architectures and methods. (a) Path Aggregation Network (PAN)) (Liu et al., 2018), (b) Reversible Columns (RevCol), (c) conventional deep supervision, and (d) our proposed Programmable Gradient Information (PGI). Three components PGI: (1) main branch (2) Auxiliary reversible branch (3) Multi-level auxiliary information

more layers. (Butakov et al., 2024)

Increasing the model size (i.e., the number of parameters) can help mitigate this issue by allowing the network to perform more complex transformations and retain more information. However, simply increasing the size does not fundamentally solve the problem of unreliable gradients in very deep networks.

### 3.2 Reversible Function

Reversible functions offer a promising solution to the problem of information loss in deep neural networks. If you apply a reversible function to data, you can reverse the process and get back the original data. This is because the nature of these functions allows the network to recover any lost information during the transformations. By incorporating reversible functions into deep neural networks, it becomes possible to reduce the problem of information loss and unreliable gradients. This offers a more effective solution compared to simply enlarging the model size.

When a function $r$ has an inverse transformation function $v$, it is called a reversible function, as shown in:

$$X = v_\zeta(r_\psi(X)),$$

where $\psi$ and $\zeta$ are parameters of $r$ and $v$, respectively. Data $X$ is converted by a reversible function without losing information, as shown in Equation:

$$I(X, X) = I(X, r_\psi(X)) = I(X, v_\zeta(r_\psi(X))).$$

When the network's transformation function is composed of reversible functions, more reliable gradients can be obtained to update the model. Most popular deep learning methods conform to this reversible property, such as:

$$X^{(l+1)} = X^{(l)} + f_{\theta^{(l+1)}}(X^{(l)}),$$

where $l$ indicates the $l$-th layer of a PreAct ResNet(He et al., 2016) and $f$ is the transformation function of the $l$-th layer. PreAct ResNet repeatedly passes the original data $X$ to subsequent layers explicitly, allowing deep networks with more than a thousand layers to converge well. However, this design can reduce the necessity of deep networks for complex tasks, explaining why PreAct ResNet performs worse than ResNet in shallower networks.

In some cases, approximation methods such as masked modeling, diffusion models, and variational autoencoders are used to find the inverse function $v$ of $r$. These methods aim to retain enough information using sparse features. However, in lightweight models, underparameterization can lead to significant loss of important information.

The information bottleneck formula highlights the challenge:

$$I(X, X) \geq I(Y, X) \geq I(Y, f_\theta(X)) \geq \cdots \geq I(Y, \hat{Y}).$$

Even though $I(Y, X)$ occupies a small part of $I(X, X)$, it is critical for the target mission. The goal is to accurately filter $I(Y, X)$ from $I(X, X)$ in lightweight models. Thus, a new deep neural network training method is proposed to generate reliable gradients suitable for shallow and lightweight networks.

## 4 Methodology implemented in Yolov9

To address the information bottleneck, YOLOv9 creators proposed a new concept, i.e., the programmable gradient information (PGI). They also designed generalized ELAN (GELAN) based on ELAN, this design of GELAN simultaneously takes into account the number of parameters, computational complexity, accuracy and inference speed. Integrating Programmable Gradient Information (PGI) and GLEAN (Generative Latent Embedding for Object Detection) architecture into YOLOv9 can enhance its performance in object detection tasks.

### 4.1 Programmable gradient information (PGI)

Programmable Gradient Information (PGI) is a solution which facilitates the generation of reliable gradients through an auxiliary reversible branch. This ensures that deep features retain the crucial characteristics necessary for executing target tasks, addressing the issue of information loss during the feed forward process in deep neural networks. It comprises a main branch for inference, an auxiliary reversible branch for reliable gradient calculation, and multi-level auxiliary information to tackle deep supervision issues effectively without adding extra inference costs.

#### 4.1.1 Main Branch Integration

The main branch of PGI optimizes the primary pathway of the network during inference, ensuring reliable and rapid processing of input data. By seamlessly integrating PGI into YOLOv9, the model benefits from enhanced gradient information and efficient layer aggregation provided by the Generalized Efficient Layer Aggregation Network (GELAN) architecture. This approach not only improves the reliability of gradient updates during training but also enhances the overall robustness and accuracy of the model in detecting objects across various scenarios and datasets.

#### 4.1.2 Auxiliary Reversible Branch

PGI introduces the auxiliary reversible branch to generate reliable gradients and update network parameters. By providing a mapping from data to targets, the loss function can guide the network and avoid finding false correlations from incomplete feed forward features that are less relevant to the target. We maintain complete information by introducing a reversible architecture. By incorporating the reversible architecture of PGI, the auxiliary branch provides additional pathways for gradient flow, ensuring more robust gradients for the loss function. This design seamlessly integrates or removes components to maintain inference speed regardless of model depth and complexity. However, combining the main branch into a reversible architecture would significantly increase inference costs. Since our goal is to use a reversible architecture to obtain reliable gradients, "reversible" is not a necessary condition during the inference stage. Thus, we consider the reversible branch as an extension of the deep supervision branch, designing the auxiliary reversible branch as shown in Figure 5(d) This gradient information drives parameter learning, assisting in extracting accurate and significant information, enabling the main branch to obtain features more effective for the target task. Additionally, the reversible architecture performs worse on shallow networks than on general networks because complex tasks require deeper network conversions. Our proposed method does not compel the main branch to retain complete original information but updates it by generating useful gradients through the auxiliary supervision mechanism. This design allows the proposed method to be applied to shallower networks as well.

#### 4.1.3 Multi-level Auxiliary Information

To resolve the challenges posed by deep supervision architectures in object detection, particularly in handling multiple prediction branches and feature pyramids for objects of different sizes, a strategic integration of multi-level auxiliary information is proposed. In Figure 5(c), the deep supervision architecture with multiple prediction branches is depicted, where each branch focuses on detecting objects at specific scales using feature pyramids.

The issue arises when shallow features are biased towards learning features necessary for small object detection, potentially treating other object sizes as background and causing critical information loss in deep feature pyramids needed for accurate predictions across all target objects.

To mitigate this, an integration network is introduced between the layers of the feature pyramid hierarchy of auxiliary supervision and the main branch, as illustrated in Figure 5(d). This network aggregates gradient information from various prediction heads, ensuring that the main branch receives comprehensive information about all target objects. By incorporating gradients containing diverse object details into parameter updates, the main branch can learn to predict effectively for various object sizes without bias or information loss.

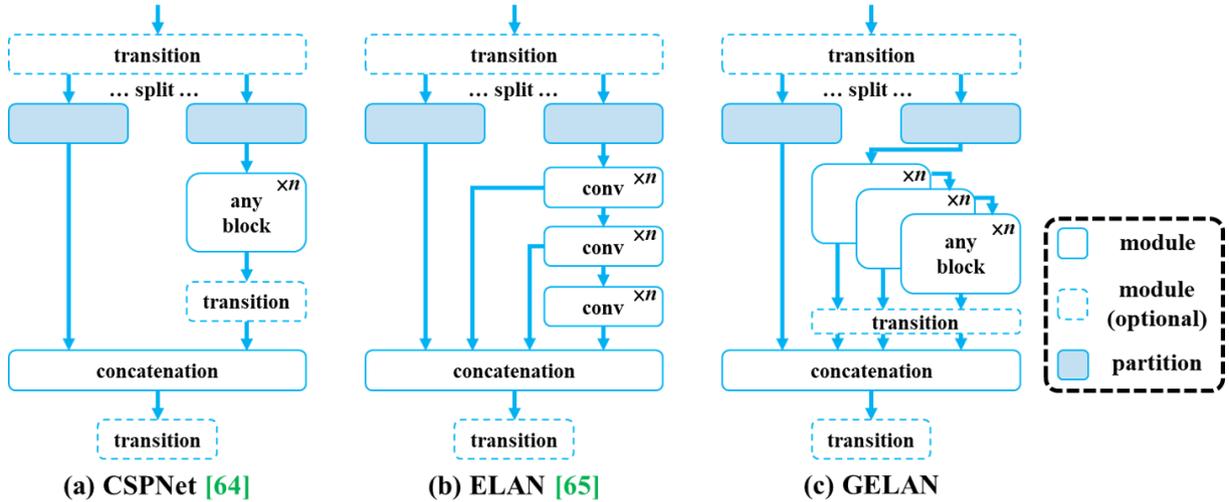

Figure 6: The architecture of GELAN: (a) CSPNet (Wang et al., 2019) , (b) ELAN(Zhang et al., 2022), and (c) proposed GELAN. We imitate CSPNet and extend ELAN into GELAN that can support any computational blocks.

This approach effectively prevents specific object information from dominating the feature pyramid hierarchy of the main branch, thereby addressing the issue of broken information in deep supervision models

### 4.2 Generalized Efficient Layer Aggregation Network (GELAN)

The integration of the Generalized Efficient Layer Aggregation Network (GELAN) into YOLOv9 significantly enhances the model's capability to process complex data patterns efficiently, supported by the Programmable Gradient Information (PGI) framework. GELAN introduces innovative layer aggregation techniques within YOLOv9, focusing on optimizing feature extraction and computational efficiency. is a combination of Spatial Pyramid Pooling (SPP) (He et al., 2014) within the ELAN structure, which starts with channel dimension adjustment through a convolutional layer. This is followed by multiple spatial pooling operations that capture contextual information across various scales. The pooled outputs are concatenated and further refined by another convolutional layer, enhancing the network's capacity to extract detailed features from diverse spatial hierarchies. It implements an advanced version of CSP-ELAN, enhancing feature extraction by splitting inputs from the initial convolutional layer into dual paths. Each path undergoes processing through RepNCSP and convolutional layers before merging, facilitating efficient gradient flow and feature reuse. This dual-path strategy optimizes learning efficiency and inference speed, maintaining depth without increasing computational complexity 's architecture merges gradient efficiency from CSPNet with ELAN's speed-oriented design, creating a unified framework adaptable to different computational environments and tasks. This flexibility empowers YOLOv9 to achieve high accuracy and rapid inference, making it suitable for a wide range of object detection applications

## 5 Architecture

The Yolov9 architecture introduces several new advancements and enhanced blocks to improve object detection accuracy and efficiency. The general and extended version of Yolov9 model use Yolov7 (Wang et al., 2022) Dynamic Yolov7 (Lin et al., 2023) respectively as a base model. The YOLO-V9 architecture for ASL object detection consists of four main components:

### 5.1 Backbone

The Backbone of YOLO-V9 is responsible for initial feature extraction from input images, progressively enhancing feature representation while main-taining computational efficiency.

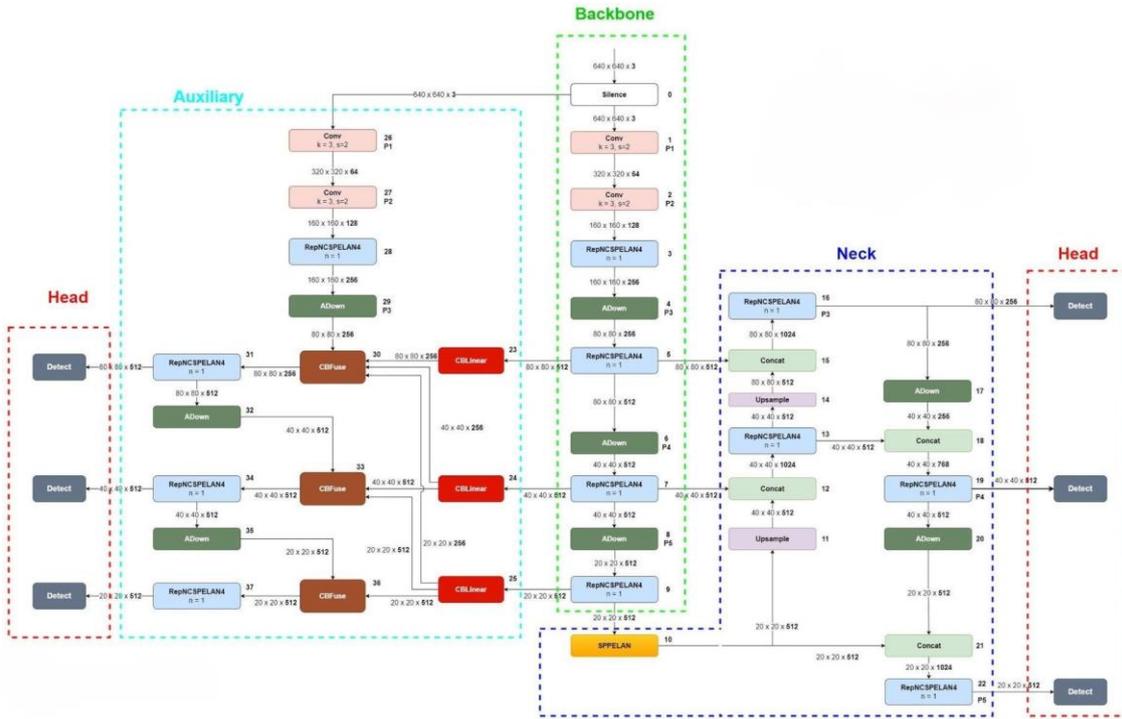

Figure 7: Working Architecture of Yolov9 model

### 5.1.1 Silence Block
This initial block efficiently passes input images to both the Auxiliary and Backbone sections without modification, ensuring seamless integration and data flow.

### 5.1.2 Convolution Blocks
Utilizing 2D convolutional layers, Batch Normalization, and the SiLU activation function, these blocks extract intricate features from the input. AutoPad dynamically adjusts padding to preserve spatial dimensions during convolutional operations.

### 5.1.3 RepNCSPELAN Block
Combining Convolutional and RepNBottleneck (Srinivas et al., 2021) blocks, this module enhances feature representation by merging multi-scale features. Inspired by ResNet bottleneck concepts, it optimizes computational resources while improving accuracy.

### 5.1.4 ADown Block
Facilitating down-sampling operations, this block splits input features for independent processing through convolutional and max-pooling layers. The results are concatenated to preserve both semantic information and spatial details.

### 5.1.5 SPPELAN Block
Employing Spatial Pyramid Pooling, this block generates fixed-size feature maps capable of robustly representing objects of varying sizes without altering input image dimensions.

## 5.2 Neck
The Neck module integrates and refines features extracted by the Backbone, preparing them for precise object detection.

### 5.2.1 Up-Sampling Layer
Increases the resolution of feature maps from the CSPELAN block to align with earlier Backbone outputs, facilitating comprehensive feature integration.

### 5.2.2 Concatenation
Merges up-sampled feature maps with corresponding resolution feature maps from earlier stages, preserving fine-grained details crucial for accurate object localization.

### 5.2.3 RepNCSPELAN Block
Further refines features through additional convolutional and RepInBotternec blocks, ensuring that aggregated features are optimized for subsequent detection tasks.

### 5.3 Auxiliary

The Auxiliary section enhances training robustness and gradient flow, critical for optimizing model performance during training phases.

#### 5.3.1 CB-Linear Blocks

These blocks generate pyramid-like feature maps across different scales, capturing high-level semantic information from the Backbone. Each CB-Linear block can output multiple feature sets with varying channel sizes.

#### 5.3.2 CB-Fuse Block

Integrates high-level features from CB-Linear blocks with lower-level features from ADown blocks, striking a balance between detailed feature representation and semantic richness. This fusion enhances model interpret ability and detection accuracy.

The Auxiliary section, operational only during training, significantly improves model convergence by providing additional gradient information and ensuring precise updates to network weights.

### 5.4 Head

The Head of YOLOv9 finalizes object detection by predicting bounding boxes and class probabilities for ASL gestures.

#### 5.4.1 Detect Blocks

Tailored for detecting ASL gestures across different scales (small, medium, large), these blocks leverage refined features from the Neck module to output precise bounding box coordinates and class predictions.

#### 5.4.2 Final Detection Output

The YOLO-V9 architecture, tailored for ASL gesture detection, delivers precise bounding box coordinates and class probabilities, offering detailed insights into detected gestures. It integrates advanced feature extraction and robust detection capabilities, leveraging components like the Auxiliary section for improved gradient flow and training robustness. YOLO-V9 excels in real-time ASL recognition with state-of-the-art performance, enhancing both accuracy and efficiency in computer vision applications, setting a strong foundation for future advancements.

### 6 Experimental Setup

Our approach includes image detection, video detection, and real-time detection, leveraging the Ultralytics (Wang et al., 2024b) platform to implement these tasks. YOLOv9 builds upon the YOLOv7(Wang et al., 2022) architecture, known for its effectiveness in various computer vision applications. Enhancements through GELAN make YOLOv9 a state-of-the-art real-time object detector. We trained two variants of YOLOv9, YOLOv9c and YOLOv9e, with 50 epochs, on a labeled dataset containing 26 ASL classes. The following sections describe the methodologies and processes for image, video, and real-time detection. Video detection extends the capabilities of image detection to sequences of frames, enabling continuous recognition of ASL signs in pre-recorded videos. This approach is particularly useful for analyzing ASL communication over time. Real-time detection applies the principles of video detection to live video streams, such as those captured from a webcam. This setup is critical for applications requiring immediate feedback, such as real-time ASL translation tools.

Image detection involves identifying ASL signs from static images using the YOLOv9 model. This task is fundamental for applications requiring precise recognition from individual frames, such as educational materials or ASL translation tools.

Through these setups, our experiment validates the effectiveness and versatility of the YOLOv9 model in detecting ASL signs across various contexts, from static images to real-time video streams. This comprehensive approach ensures that the model can be reliably used in practical applications, enhancing communication tools for the deaf and hard-of-hearing community.

### 7 Results

The mAP50-95 metric comparison of both modals YOLO-v9e and YOLO-v9c is given in Figure 8 below. mAP defines the model's ability to correctly identify and localize objects in an image. Here, After 50 Epochs with Batch size 10 YOLO-v9e is at $69.38\%$ and YOLO-v9c is at $68.56\%$. The YOLO-v9e model achieved 96.83% precision, 92.96% recall, and 97.84% mAP @0.5, whereas, The YOLO-v9c model achieved 93.84% precision, 90.44% recall, and 96.46% mAP50. Showing the model capabilities in real-time hand gesture recognition.

The Figure 12 and Figure 13 show the real-time detection results for both YOLOv9c and YOLOv9e models. Each gesture is correctly identified and bounded by colored boxes corresponding to the gesture labels. From the table 14, it is evident that YOLOv9c is significantly faster in detecting ASL gestures compared to YOLOv9e. The average detection times for YOLOv9c are approximately half of those for YOLOv9e across all gesture categories.

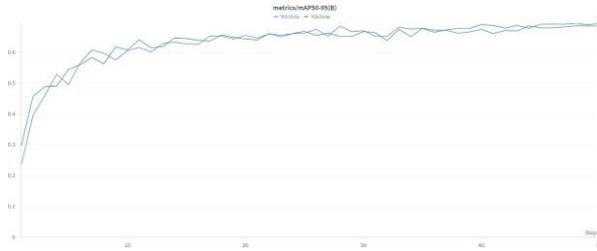

Figure 8: Metric/mAP50-95 comparison between YOLO-v9e and YOLO-v9c for Batch Siye = 10 and Epochs = 50.

The bounding box loss in YOLO-v9c and YOLO-v9e models is 0.324 and 0.3088, while the classification loss of YOLO-v9c and YOLO-v9e models is 0.1905 and 0.1772. This shows that YOLO-9e is better at detecting objects and has better classification accuracy. Overall, v9e is more accurate and faster at detecting objects and recognizing hand gestures.

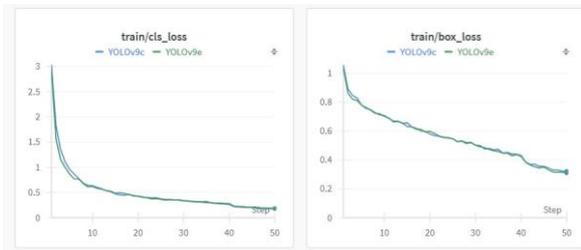

Figure 9: Cls-loss and Box-loss comparison between YOLO-v9e and YOLO-v9c for Batch Size = 10 and Epochs = 50.

After training, Both the model YOLO-v9c and YOLO-v9e were tested on an image with all 26 letters in it. Figure 10 shows the test result of YOLO-v9c model and Figure 11 shows the test result of YOLO-v9e model. Both the model we able to detect all 26 letters in the image correctly.

## 8 Conclusion

YOLO is the state-of-the-art computer vision model for classification and object detection. It has evolved impressively over a decade, making further possibilities in object detection. Due to its speed and accuracy, it is a great option for real-time object detection.

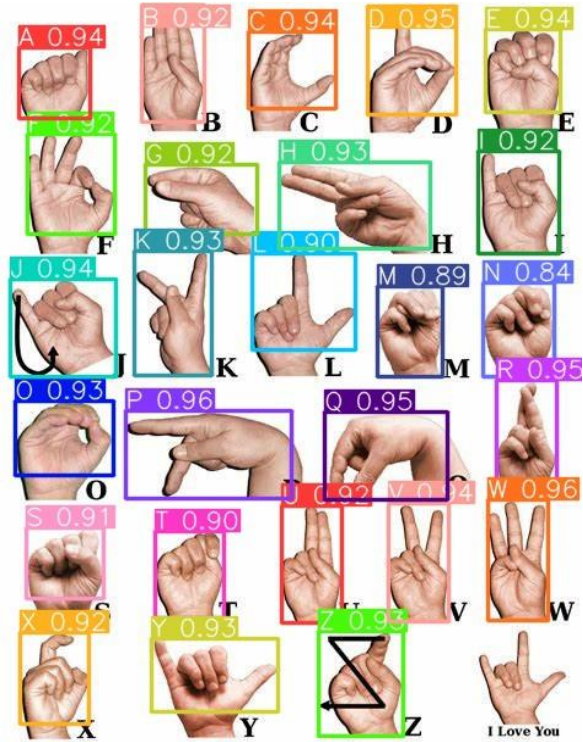

Figure 10: YOLO-v9c result on multiple ASL sign detection

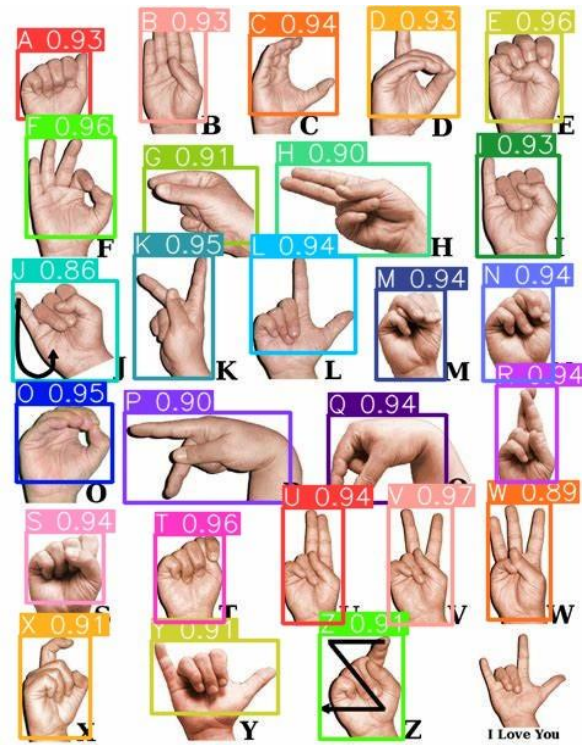

Figure 11: YOLO-v9c result on multiple ASL sign detection

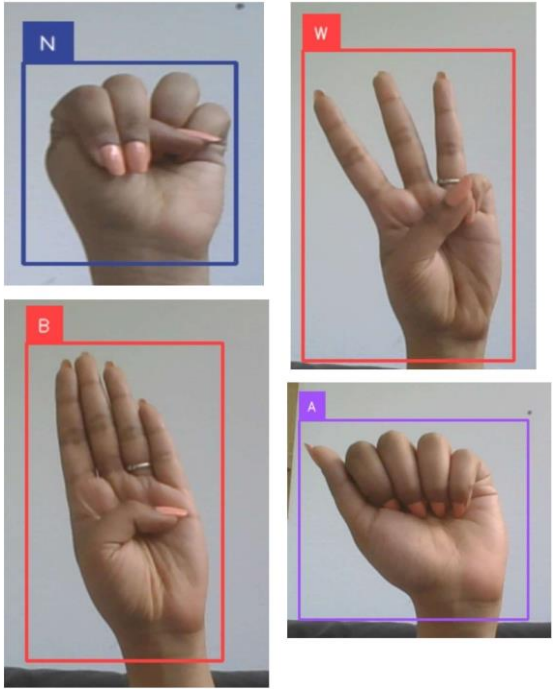

Figure 12: Real-time detection using YOLO-v9c on multiple ASL signs

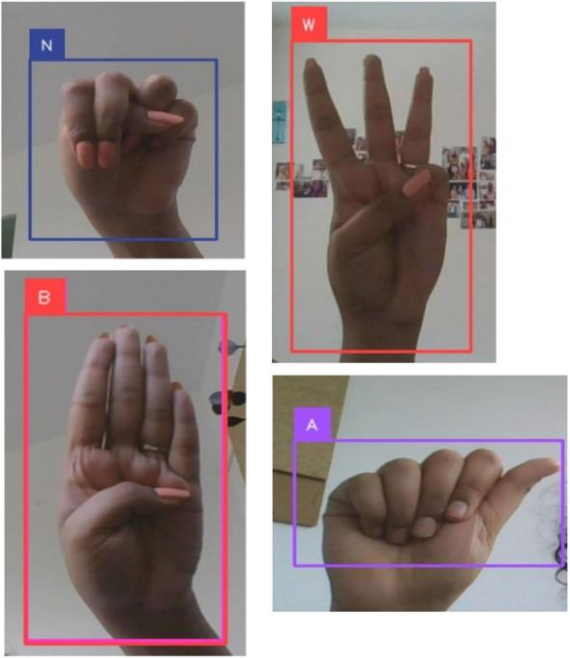

Figure 13: Real-time detection using YOLO-v9e on multiple ASL signs

| Model | Time taken to detect | | | |
|---|---|---|---|---|
| Yolov9c | 1349.7ms(N) | 1371.3ms(B) | 1358.1ms(A) | 1344.5ms(W) |
| Yolov9e | 2650.9ms(N) | 2728.9ms(B) | 2852.6ms(A) | 2776.8ms(W) |

Figure 14: Table shows the time taken for our models to detect ASL in real time

The YOLOv9c model demonstrates superior performance in terms of detection speed when compared to the YOLOv9e model. This efficiency makes YOLOv9c more suitable for applications requiring real-time ASL gesture recognition. The visual results also confirm that both models accurately detect the specified gestures, although the faster response time of YOLOv9c may offer an advantage in dynamic, real-time environments. PGI eliminate the information bottleneck problem in earlier models making detection more efficient and faster, allowing new lightweight architectures to be applied in daily life, whereas GELAN uses conventional convolution to achieve a higher parameter usage, showing the advantages of being lightweight, fast, and accurate. The new YOLO-v9, designed by combining PGI and GELAN, provides the opportunity to integrate real time sign language detection with smarts gadgets.

### 8.1 Future Scope

The application of the YOLOv9 model for American Sign Language (ASL) detection represents a significant advancement in computer vision, particularly in accessibility and communication technologies. Looking ahead, there are several promising avenues for future development and enhancement:

1. **Real-Time Sign-to-Speech Translation**: Future advancements could integrate ASL recognition with speech synthesis technologies, enabling real-time translation of ASL signs into spoken language.

2. **Multi-modal Integration**: Future models could incorporate additional sensory inputs like depth sensing or spatial awareness to improve ASL gesture interpretation in varied environments.

3. **Ethical Considerations**: Ongoing attention to ethical considerations, including privacy, data security, and cultural sensitivity, is essential in the development and deployment of ASL detection systems.

In conclusion, the future of ASL detection and translation shows great potential to enhance accessibility and communication for sign language users. Through ongoing technological advancements and collaboration, integrating ASL detection with speech synthesis marks a significant stride towards inclusive and empowering communication tools for the deaf and hard-of-hearing community.